\documentclass{article}

\usepackage{arxiv}
\usepackage{tikz} 
\usetikzlibrary{calc}
\usepackage{enumitem}
\usepackage{amssymb}
\usepackage{comment}
\usepackage[utf8]{inputenc} 
\usepackage[T1]{fontenc}    
\usepackage{hyperref}       
\usepackage{url}            
\usepackage{booktabs}       
\usepackage{amsfonts}       
\usepackage{nicefrac}       
\usepackage{microtype}      
\usepackage{lipsum}		
\usepackage{graphicx}
\usepackage{natbib}
\usepackage{amsmath}
\usepackage{xcolor}

\usepackage{doi}

\title{One target to align them all:\\ LiDAR, RGB and event cameras extrinsic calibration for Autonomous Driving}

\author{ \href{https://orcid.org/0009-0004-7287-891X}{\includegraphics[scale=0.06]{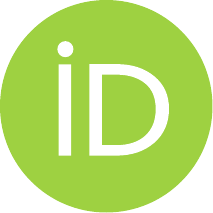}\hspace{1mm}Andrea Bertogalli}\thanks{Corresponding author: \texttt{andrea.bertogalli@mail.polimi.it}} \\
	DEIB\\
	Politecnico di Milano\\
	Milan, IT \\
	\texttt{andrea.bertogalli@mail.polimi.it} \\
	\And
	\href{https://orcid.org/0000-0002-1650-3054}{\includegraphics[scale=0.06]{orcid.pdf}\hspace{1mm}Giacomo Boracchi} \\
	DEIB\\
	Politecnico di Milano\\
	Milan, IT \\
	\texttt{giacomo.boracchi@polimi.it} \\
    	\And
	\href{https://orcid.org/0000-0002-0598-8279}{\includegraphics[scale=0.06]{orcid.pdf}\hspace{1mm}Luca Magri} \\
	DEIB\\
	Politecnico di Milano\\
	Milan, IT \\
	\texttt{luca.magri@polimi.it} \\
}

\newcommand{\blackletter}[1]{%
  \tikz[baseline=(char.base)]{
    \node[shape=circle,draw=black,fill=black,inner sep=1pt] (char)
    {\textcolor{white}{\textbf{#1}}};
  }%
}

\def\etal{\emph{et al. }}

\hypersetup{
pdftitle={A template for the arxiv style},
pdfsubject={q-bio.NC, q-bio.QM},
pdfauthor={David S.~Hippocampus, Elias D.~Striatum},
pdfkeywords={First keyword, Second keyword, More},
}

\begin{document}
\maketitle

\begin{abstract}
	We present a novel \emph{multi-modal extrinsic calibration} framework designed to simultaneously estimate the relative poses between event cameras, LiDARs, and RGB cameras, with particular focus on the challenging event camera calibration. Core of our approach is a novel 3D calibration target, specifically designed and constructed to be concurrently perceived by all three sensing modalities. The target encodes features in planes, ChArUco, and active LED patterns -- each tailored to the unique characteristics of LiDARs, RGB cameras, and event cameras respectively. This unique design enables a one-shot, joint extrinsic calibration process, in contrast to existing approaches that typically rely on separate, pairwise calibrations.
Our  calibration pipeline is designed to accurately calibrate complex vision systems in the context of autonomous driving, where precise multi-sensor alignment is critical. We validate our approach through an extensive experimental evaluation on a custom built dataset, recorded with an advanced autonomous driving sensor setup,  confirming the accuracy and robustness of our method. Further implementation details are available online at the  \href{https://andberto.github.io/One-target-to-align-them-all/}{supplementary material page}.
\end{abstract}

\section{Introduction}
\label{sec:intro}

In autonomous driving, a diverse set of sensing modalities is required to gain a complete understanding of the environment. RGB cameras provide semantic information, LiDARs offer precise 3D geometric structure, and event cameras are particularly effective in challenging conditions such as high-speed motion, low light or high dynamic range scenes. While these sensors are complementary, their integration relies critically on accurate \emph{extrinsic calibration}, \emph{i.e.}, the estimation of the relative poses between each pair of sensors in the system. Among these sensor modalities, event cameras present the most significant calibration challenges due to their fundamentally different sensing mechanism and sparse, asynchronous output. While LiDAR-RGB calibration is well-established in the literature, robust event camera calibration, particularly in multi-sensor contexts, remains an open problem that our work specifically addresses.

\begin{figure}
    \centering
    \includegraphics[width=1\linewidth]{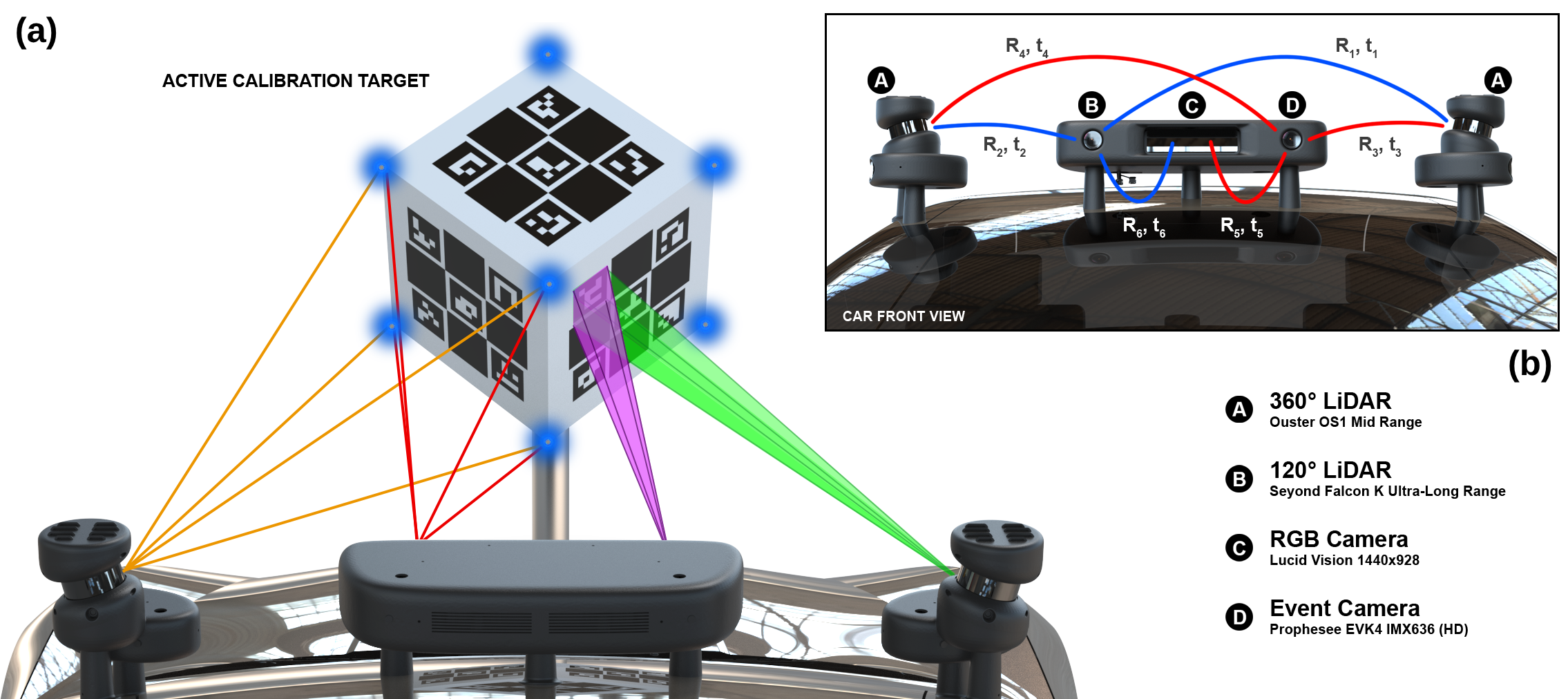}
    \caption{An overview of our calibration target.
        (a) Our target allows simultaneous features detection across all reference frames. Specifically, the target design addresses event camera detection challenges through frequency-coded LEDs placed at the corners of the cube.
        (b) The sensor suite along with the estimated relative poses $(R_{ij}, t_{ij})$ between each sensor pair $(i, j)$. Note that not all RGB cameras are visualized for clarity.} 
    \label{fig:cube}
\end{figure}

Specifically, as illustrated in Figure~\ref{fig:cube} (b), extrinsic calibration  defines the spatial transformation that aligns the coordinate systems of different sensors, enabling consistent sensors fusion and coherent scene reconstruction. The goal is to compute the rotation matrices $R_{ij}$ and translation vectors $\mathbf{t}_{ij}$	
  that relate the coordinate systems of each sensors' pair $(i,j)$. 
  Note that, unless otherwise specified, the term calibration refers exclusively to the estimation of extrinsic parameters, thus the relative pose of the sensors, as we assume intrinsic parameters already estimated by classical approaches for each sensor. 
  
  We focus on the estimation of the relative pose of a suite of \emph{heterogeneous sensors} including an event camera, LIDARs and  RGB cameras. This is essential in multi-sensor frameworks such as those used in autonomous driving for object detection \cite{sgaravatti2025multimodal, bai2022transfusion, yin2023fgfusion}, but also applies  to other robotics and perception tasks as odometry estimation \cite{censi2014low, zuo2019lic}, drone tracking \cite{magrini2024neuromorphic}, obstacle detection \cite{wu2023flytracker} and a broad range of additional use cases.
  
 Unfortunately, this task is inherently challenging, as the diverse sensor modalities make it non-trivial to establish unambiguous and reliable correspondences, since there is no direct one-to-one mapping between features of each modality.
In particular, establishing correspondences between the sparse events returned by an event camera and the dense LiDAR pointcloud remains a significant challenge.
The calibration is further complicated by the high precision required to ensure an accurate spatial alignment of the sensors. 
These characteristics make traditional calibration techniques less effective. Moreover, classical calibration methods are not suited for event cameras, as they address completely different data. Most of the few methods that address the calibration of an event camera with a LiDAR rely on double-projection schemes \cite{zhu2018multivehicle, gehrig2021dsec}.  These approaches, instead of calibrating the event camera directly with the LiDAR, follow a two-step approach:  the event camera is calibrated with an auxiliary sensor (tipically an RGB camera), which is in turn aligned with the  LiDAR. This approach suffers from reduced robustness and precision, as the final accuracy depends on the compounded error of multiple transformations, and often requires multiple calibration rigs.
In contrast, our method overcomes these limitations by providing a \emph{single}, \emph{direct} estimate of all relative poses \emph{simultaneously} during a single data acquisition. 

To achieve this one-shot multi-modal calibration, we introduce a custom 3D calibration target comprising a three-faced ArUco cube with blinking LEDs on each corner as shown in Figure~\ref{fig:cube} (a). Our target is specifically designed to provide salient features across all three sensor modalities. In addition, our calibration scheme includes a novel feature detection approach for both LiDAR and events in order to provide an accurate and simultaneous feature matching between the two. 
Our method is well suited for autonomous driving scenarios where the intrinsic parameters of each sensor are typically constant over time and the calibration scene is  static.  Furthermore, temporal synchronization issues, that are common in other methods due to the different sampling rates of the involved sensors, are inherently avoided by design. As demonstrated in our experiments,  our method enables precise and unified extrinsic calibration across multiple modalities, showing improved accuracy and robustness compared to  state-of-the-art alternatives \cite{song2018calibration, xing2023target, yan2023joint}.

To our knowledge, no existing work has proposed a pipeline capable of directly calibrating a LiDAR with both event and RGB cameras simultaneously. Thus in the following section we focus on works that address the calibration of pairs of sensors.

\section{Related works}
\label{sec:related_works}

Given the relatively recent adoption of event cameras, the body of literature on multi-sensor calibration involving them is still rather limited. In the following, we briefly review the main existing approaches that address  calibrations between pairs of different sensor modalities, considering both target-based and targetless approaches and also automatic and manual approaches. In practice, target-based calibration methods are often preferred due to their higher accuracy, repeatability, and ease of use, especially in controlled environments, unlike target-free approaches, which typically rely on motion or scene assumptions that may not hold consistently.

\paragraph{LiDAR - event camera extrinsic calibration}
 The literature on extrinsic calibration between an event camera and a LiDAR is  scarce. Most datasets \cite{gehrig2021dsec, zhu2018multivehicle} use the double projection scheme to project points between the event camera's reference frame and the LiDAR's one, passing through the reference frame of an RGB camera. This approach is simpler than direct calibration but more imprecise, as it compounds the errors of two projections.

There are only few direct calibration methods between LiDARs and event cameras. In Ta \etal \cite{ta2023l2e} the authors exploit the fact that most recent event cameras can capture the reflection of the laser beams projected by a LiDAR sensor, and this is used to perform an automatic target-free calibration by minimizing an entropy loss. Jiao \etal \cite{jiao2023lce} proposes a calibration  based on deep-frame reconstruction with the well-known E2VID model \cite{rebecq2019high}. Another automatic target-free approach is  presented by Xing \etal \cite{xing2023target} which exploits motion to generate events and calibrate the sensors. The only method directly comparable to ours is that of Song \etal \cite{song2018calibration}, which introduces an automatic target-based event camera–LiDAR calibration using a custom calibration board with four illuminated circular holes. While these circular patterns can be captured by both the LiDAR and the event camera, a different target is required to calibrate the RGB camera with the LiDAR, and the resulting performance is inferior to that of our multi-modal approach.

\paragraph{LiDAR - RGB camera calibration}
Extrinsic calibration between LiDARs and RGB cameras is a well explored task \cite{li2023automatic, survey2}, with existing approaches generally falling into four main categories. A first group includes \emph{manual target-based methods}, where a user manually identifies correspondences on a known calibration target before performing the calibration. These methods commonly use checkerboard patterns \cite{geiger2012automatic, zhang2004extrinsic, zhou2012new}, ArUco markers \cite{dhall2017lidar, yoo2018improved}, or other types of fiducial points \cite{guindel2017automatic, velas2014calibration, hassanein2016new, pusztai2017accurate}.

A second group consists of \emph{automatic target-based methods}, where no manual annotation is required, making them particularly well-suited for real-world applications in controlled environments. For example, \cite{toth2020automatic, beltran2022automatic, yan2023joint, grammatikopoulos2022effective, zhou2018automatic} propose fully automatic calibration pipelines based on designed targets. In particular, Yan \etal \cite{yan2023joint} introduce a custom target composed of circular holes and a checkerboard, enabling reliable feature detection in both LiDAR and RGB modalities—a design also adopted by \cite{song2018calibration} for event-LiDAR calibration. Grammatikopoulos \etal \cite{grammatikopoulos2022effective} propose a target which exploits two crossing retroreflective
stripes and an AprilTag to calibrate a LiDAR and a camera, but it requires a dynamic scene and a consequent temporal calibration between the two sensors. Similarly Zhou \etal \cite{zhou2018automatic} employ an acquisition of a classic checkerboard, exploiting lines and planes correspondences to calibrate the sensors.

A third class involves \emph{manual targetless methods}, where correspondences are manually selected directly between RGB images and LiDAR pointclouds \cite{pervsic2021online}. While flexible, these approaches are typically labor-intensive and less repeatable. Lastly, \emph{automatic targetless methods} aim to infer extrinsic parameters directly from environmental features such as planes or edges, relying on assumptions about the scene's structure or motion \cite{li2017online, liu2018deep, yuan2021pixel}. Although promising for online or in-field calibration, these methods often fall short in terms of precision and robustness compared to target-based alternatives.
\paragraph{RGB camera - event camera extrinsic calibration}
The most common approach for calibrating an event camera with an RGB camera is to apply standard stereo calibration techniques. The is achieved by reconstructing a temporally synchronized frame-like representation of the event stream either with deep learning \cite{rebecq2019high, muglikar2021calibrate} or with other techniques \cite{hu2024dynamic, dubeau2020rgb, cress2024tumtraf}. This reconstruction provides two synchronized data streams, from which features are detected and matched to perform stereo calibration.

\begin{figure}[t]
    \centering
    \includegraphics[width=1\linewidth]{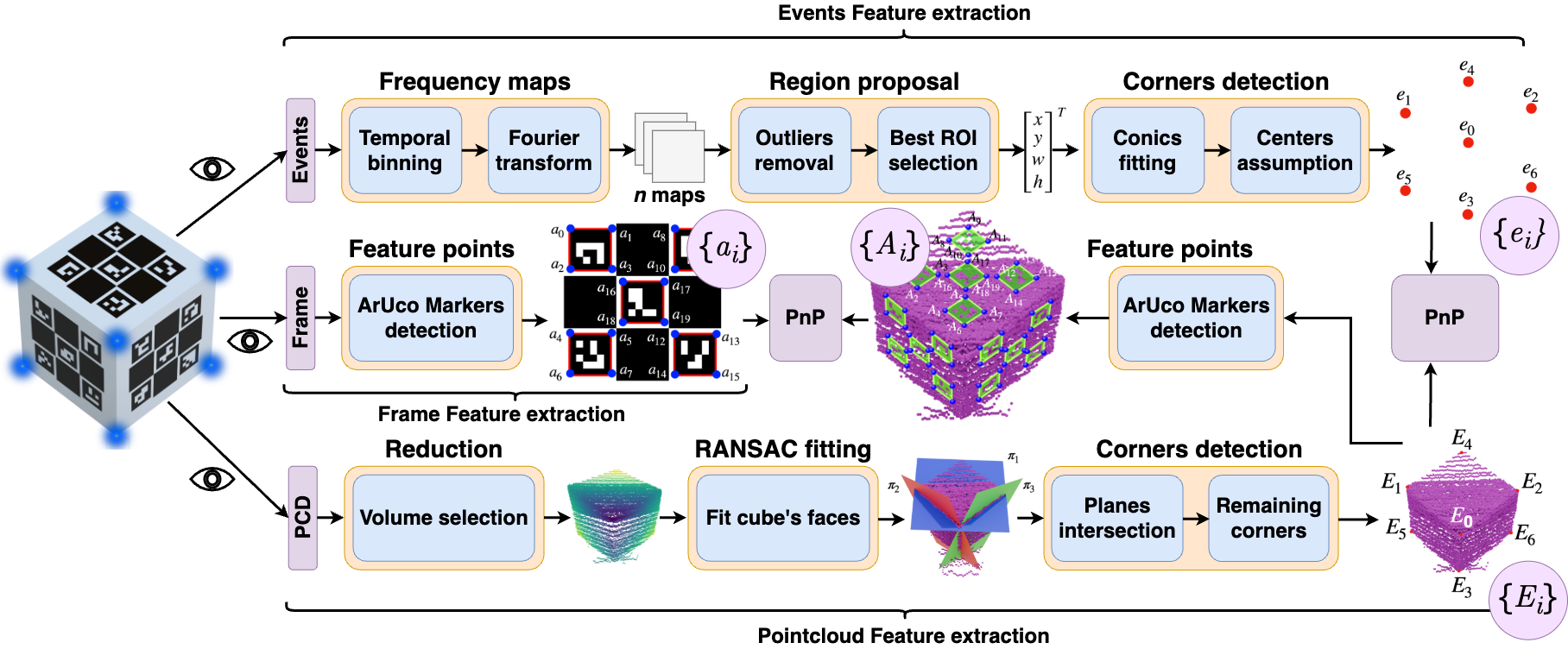}
    \caption{Our calibration pipeline consists in three feature detection branches on the same calibration target. Our novel event feature detection branch detects the 7 cube's corners $e_{i}$, the RGB feature detection detects the ArUcos' corners $a_{i}$ and the pointcloud feature detection branch detects the 3D points corresponding to cube's corner $E_{i}$ and ArUcos' corners $A_i$. Finally we apply the PnP algorithm on the correspondences $A_i \leftrightarrow a_i$ and $E_i \leftrightarrow e_i$.}
    \label{fig:pipeline}
\end{figure}

\section{Method}
\label{sec:method}
The core idea of our method is to leverage a custom-designed calibration target that interacts with all sensing modalities and enables the \emph{simultaneous} detection of the same physical keypoints across the different reference frames involved. Our target consists in a cube with blinking LEDs at its corners, as illustrated in Figure~\ref{fig:cube} and described in Section~\ref{sec:target}. Our calibration framework is structured into three main keypoints extraction modules, one for each modality involved, as illustrated in Figure~\ref{fig:pipeline}. 
The core innovation lies in detecting blinking LEDs placed on the cube's corners $\{E_i\}$ to extract keypoints $\{e_i\}$ from the event stream through frequency analysis (Section~\ref{sec:corners_events}).  The  three orthogonal faces of the cube are exploited to extract the corresponding corners $\{E_i\}$ in the LiDAR pointcloud (Section~\ref{sec:lidfeat}), while ArUco markers $\{A_i\}$ provide anchor points $\{a_i\}$ for RGB cameras (Section~\ref{sec:feat_RGB}). Crucially, the physical layout of the target ensures that the extracted features are not only detectable, but also uniquely matched across modalities. This allows us to compute correspondences $\{e_i\leftrightarrow E_i\}$ and $\{a_i\leftrightarrow A_i\}$ without ambiguity. Once these cross-sensor correspondences are established, we estimate the extrinsic transformations using a standard Perspective-n-Point (PnP) formulation (Section \ref{sec:pose_est}). 

\subsection{3D multi-modal target}
\label{sec:target}

Our custom target design is specifically optimized for event camera feature detection challenges. We adapt the blinking LED pattern to easily detect unique keypoints and establish reliable correspondences between sparse event data and dense LiDARs pointcloud.

\begin{figure}[tb]
    \centering
    \includegraphics[width=0.9\linewidth]{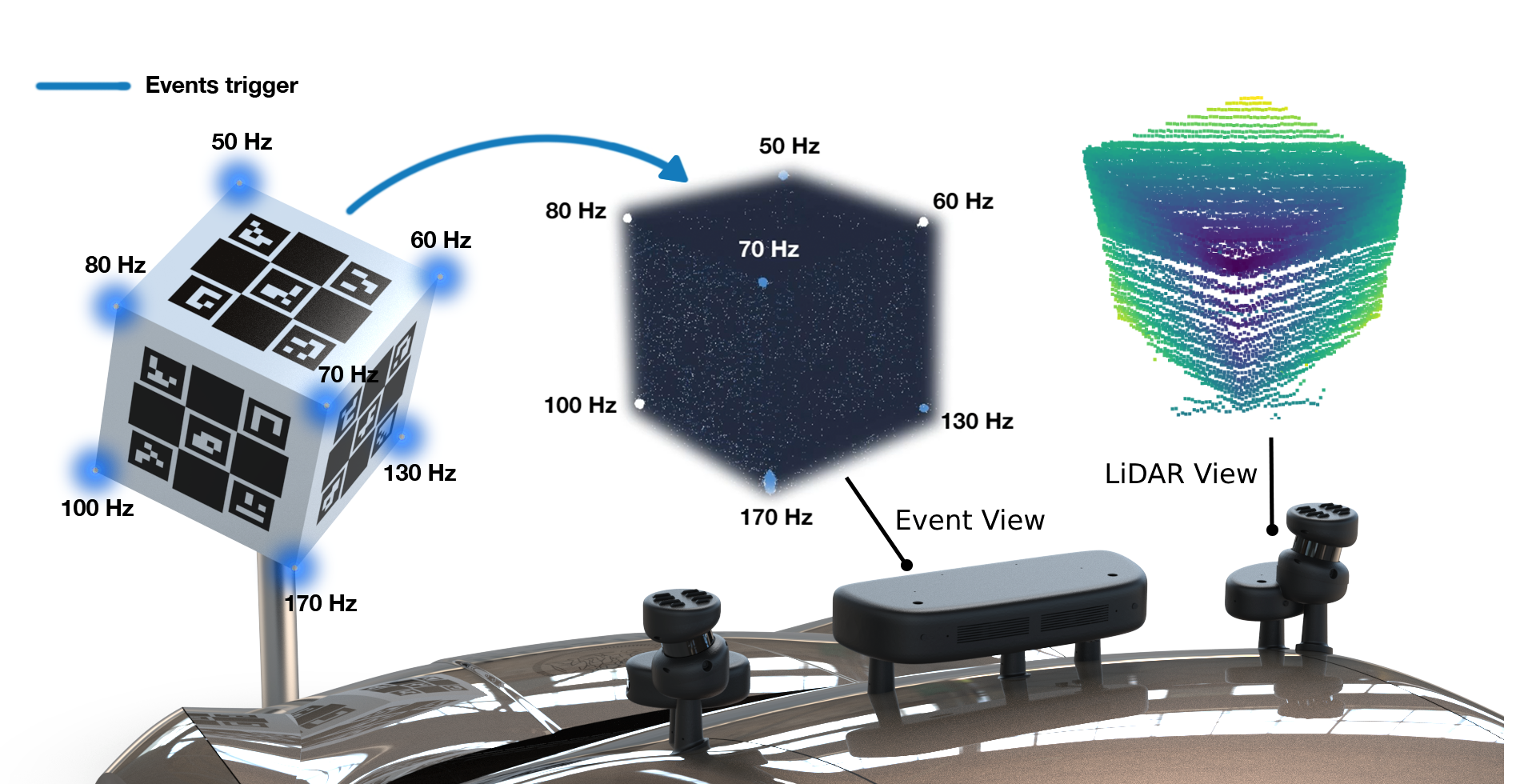}
    \caption{The operating principle of our calibration target. Each LED on the cube blinks at a specific frequency, resulting in a continuous event generation on the event camera yielding a square wave signal for each pixel. In the event view is shown a cumulated frame-like representation of the event stream over a \textit{33.333 ms} time period. The LiDAR view emphasizes the cube inside the pointcloud, where darker colors represent points that are closer to the sensor.}
    \label{fig:maserati_trigger}
\end{figure}

We only consider the three adjacent faces of the cube facing the sensor suite, that are made by assembling together 3 PVC planes.
On each face we place a different $3\times3$ ChArUco board  to uniquely identify the markers' features $\{a_0,a_1,...,a_{59}\}$ from the RGB camera, while the 3D geometry of the object allows to identify the corners $\{E_0,E_1,...,E_6\}$ and Aruco's 3D positions $\{A_0,A_1,...,A_{59}\}$ in the same reference system of the pointcloud. On the seven visible corners $\{E_i\}$ of the 3 faces we installed 7 ellipsoidal LED diodes. 
LEDs are controlled by an ESP32 microcontroller which makes them blink at a specific frequency (reported in Figure \ref{fig:maserati_trigger}). Frequencies have been chosen  in order to mitigate the risk of ambiguity in uniquely identifying the LED in the subsequent analysis. The blinking behavior is essential  to trigger events while the blinking frequency is used to uniquely identify each LED. Each LED represents a feature in the event stream $\{e_0,e_1,...,e_6\}$. Note that the measures of the cube are known, allowing a precise modeling of the target.

\subsection{Identifying  keypoints in the event stream}
\label{sec:corners_events}
The procedure to identify the position of the keypoints in the event stream is composed by three main stages.

\begin{figure}[t]
\centering
\begin{tabular}{ccc}
\includegraphics[height=2.65cm]{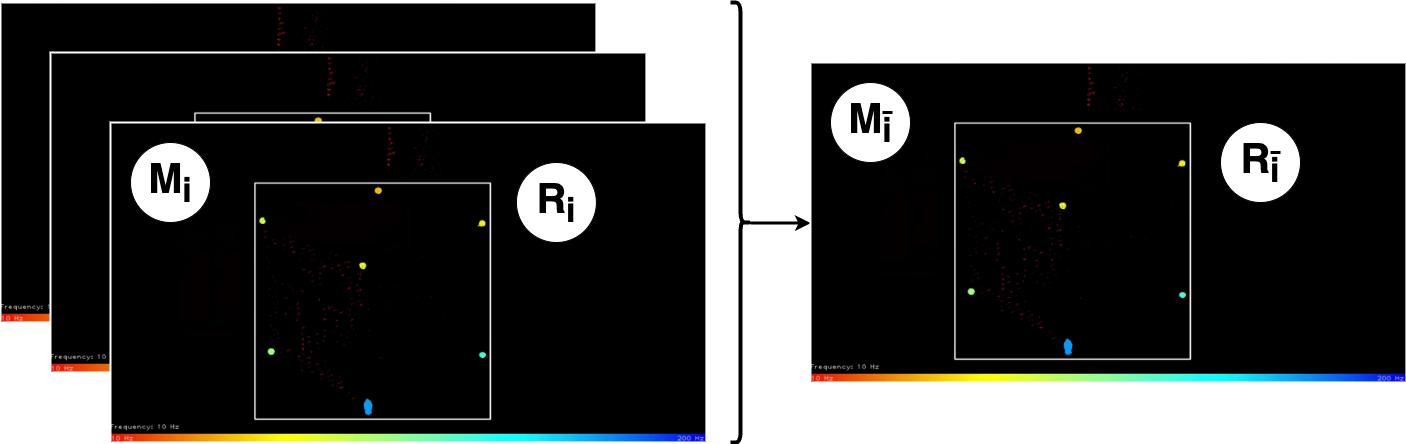} &
\includegraphics[height=2.65cm]{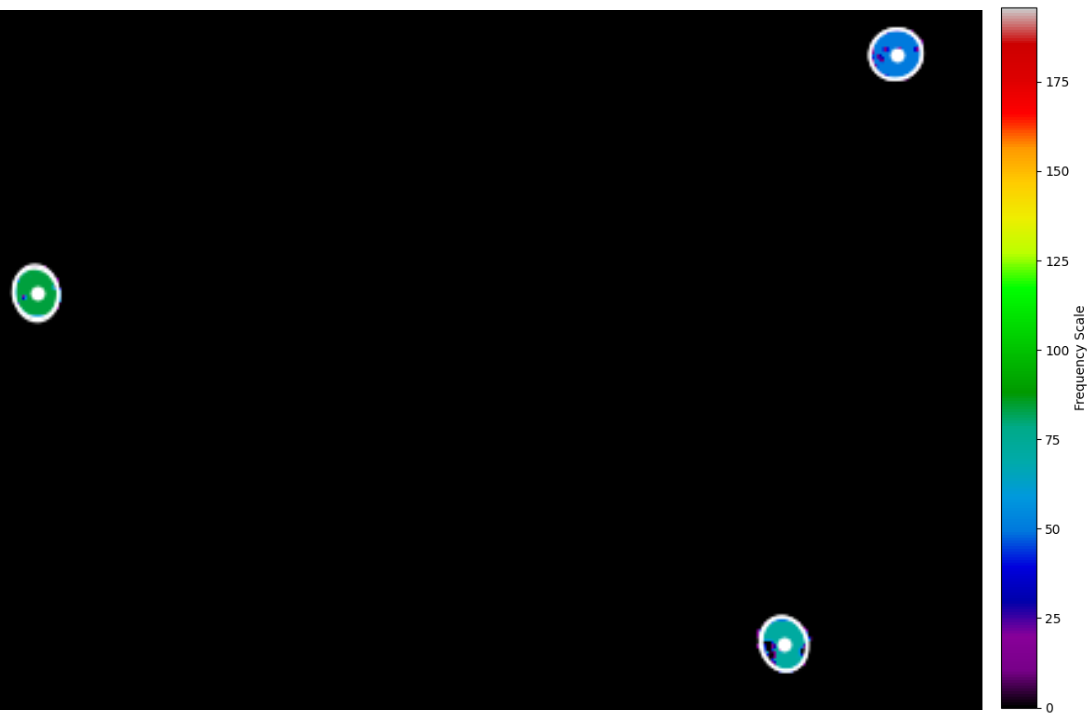} \\
(a) & (b)
\end{tabular}
\caption{On the left: $n$ frequency maps $\mathbf{M}_{i}$ with their bounding boxes $R_i$  are analyzed to yield the best frequency map $\mathbf{M}_{\bar{i}}$ along with the bounding box $R_{\bar{i}}$. On the right, the figure shows some of the detected points (the centers of the fitted ellipses) $e_i$, which are identified using the associated frequency. Note that only three ellipses are shown for simplicity.}
\label{fig:ev_features}
\end{figure}

\noindent\textbf{i) Frequency analysis:} 
Pixels of an event camera respond to changes in logarithmic brightness intensity \( L = \log(I) \). An event is a tuple \( \eta_k = (u_k, t_k, p_k) \), where \( u_k = (x_k, y_k)^T \) is the pixel location, \( t_k \) the timestamp (typically with a temporal resolution of around $10~\mu\mathrm{s}$), and \( p_k \in \{-1, +1\} \) the polarity. In particular, an event is triggered when the change in intensity \( L \) exceeds a threshold \( C \) since the last event at \( u_k \), with polarity:  
\begin{equation}
p_k = \begin{cases} 
+1, & L(u_k, t_k) - L(u_k, t_k - \Delta t) \geq C \\ 
-1, & L(u_k, t_k) - L(u_k, t_k - \Delta t) \leq -C,\\
\end{cases}
\end{equation}

\noindent where \( \Delta t = t_k - t_{k-1} \) is the elapsed time between two consecutive events. Note that pixels with small intensity variation do not give rise to any event.
 When the LEDs on the target turn on and off repeatedly each pixel generates a stream of events that can be interpreted as a periodic square wave signal. Given an event stream we can compute the blinking frequency of each blinking LED by means of a \emph{frequency map}, defined as a matrix $\mathbf{M} \in \mathbb{R}^{w \times h \times 1}$ having the same dimension $(w,\;h)$  of the event camera frame. A frequency map stores for each pixel its blinking frequency value in $Hz$ as depicted in Figure \ref{fig:ev_features} (a), where frequencies are color-coded. 
 Computing frequencies from an event stream is an already explored task \cite{zhao2024power, pfrommer2022frequency, hoseini2017passive, censi2013low}, in our implementation we adopt a method similar to \cite{zhao2024power}.
 The frequency detection is limited on the predefined range of frequencies (10-200hz), acting like a band-pass filter to filter out environmental noise. 

In order to construct a frequency map $\mathbf{M}$ we compute $m$ temporal bins of length $\Delta t$:
\begin{equation}
    B_j = \left[ t_0 + (j - 1) \Delta t, t_0 + j \Delta t \right)\;\;\text{for\;\;} j = 1, 2, \dots, m.
\end{equation}
 For each bin, we derive a value of the signal by considering the polarity of the events in the bin. Once we have reconstructed the square wave for each pixel, we  apply the Fast Fourier Transform  to obtain the frequency map and consequently the blinking frequency of the LEDs.

\noindent\textbf{ii) Best frequency map estimation:}
During the acquisition, it is  very likely that interferences arise in the event stream and spoil the resulting frequency map. Thus, rather than considering a single frequency map for the entire stream, we compute $n$ frequency maps $\mathbf{M}_i$, each one for a portion of the stream. This allows us to strengthen the frequency estimation process.
On each $\mathbf{M}_i$ we compute the smallest bounding box $R_i$ that encloses the non-null frequency region, depicted in white in Figure~\ref{fig:ev_features} (a).
Then we discard all the bounding boxes that are deemed outliers according to the 3-$\sigma$ rule. 
Eventually, we compute a reference bounding box $R$ having as vertices the means of the corresponding vertices of all the remaining $R_{i}$. We then select the frequency map $\mathbf{M}_{\bar{i}}$ whose bounding box $R_{\bar{i}}$ minimizes the Manhattan distance ($L_1\text{-norm}$) with $R$:
\begin{equation}
\bar{i} = \arg \min_j \|R_j - R\|_{1}.
\end{equation}


\noindent\textbf{iii) Ellipses fitting and location estimation:} As last step, inside the selected bounding box $R_{\bar{i}}$ we fit in $\mathbf{M}_{\bar{i}}$ seven ellipses, one for each LED, using a constrained least squares \cite{fitzgibbon1999direct} on the edges \cite{canny} detected in the frequency map. At this point, we assume that each ellipse's center $\{e_0, e_1, ...,
e_6\}$ corresponds to a cube's corner as illustrated in Figure \ref{fig:ev_features} (b).

\subsection{Identifying 3D corners on the pointcloud}
\label{sec:lidfeat}

\begin{figure}
\centering
\begin{tabular}{ccc}
\includegraphics[height=2.9cm]{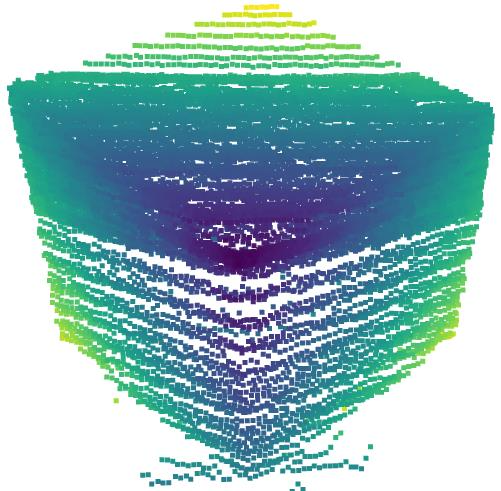} &
\includegraphics[height=2.9cm]{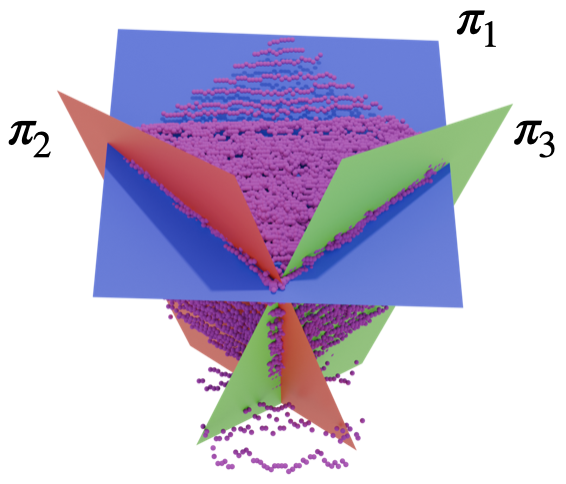} &
\includegraphics[height=2.9cm]{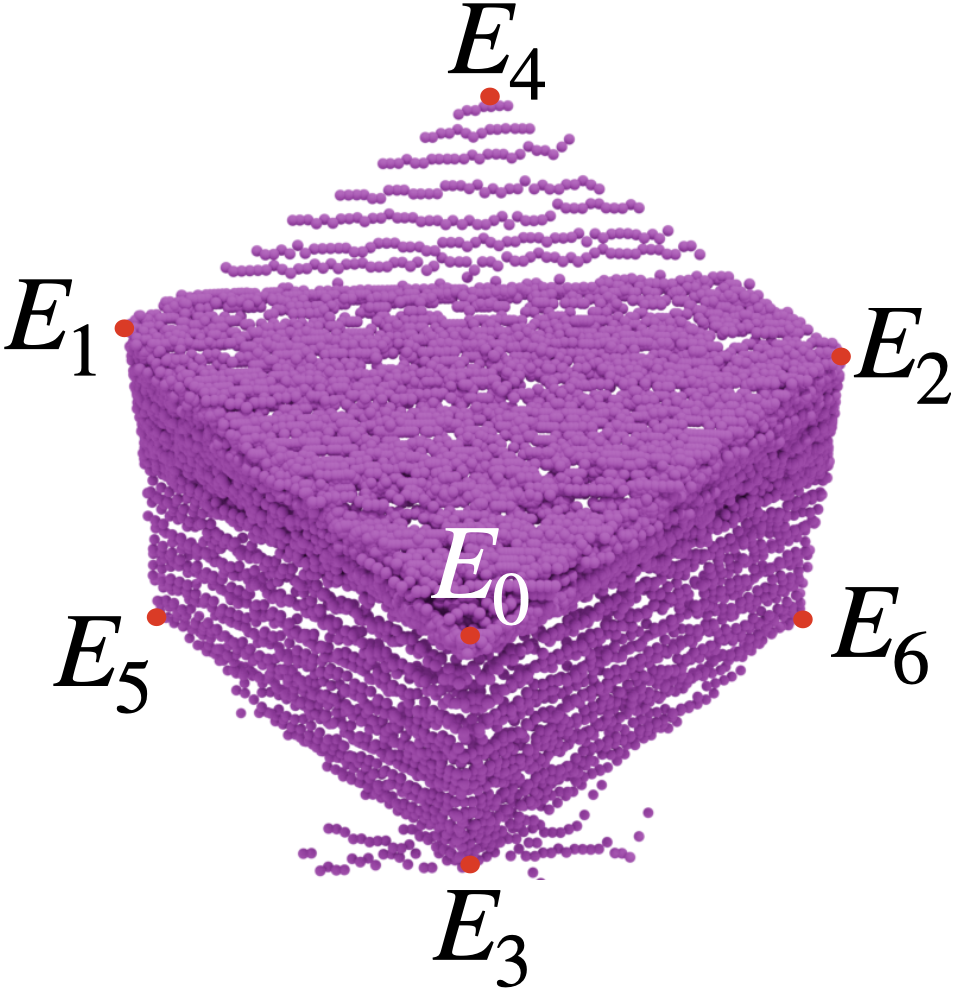} \\
(a) & (b) & (c)
\end{tabular}
\caption{The output of the three steps to detect the features (the seven cube's corner $E_i$) from the LiDAR pointcloud. (a) the reduced pointcloud we are considering. (b)  the fitted planes which represent the cube's faces. (c)  the seven detected points $E_i$, which are known exploiting the geometry of the cube.}
\label{fig:lid_pipeline}
\end{figure}

To identify  the 3D positions $\{E_i\}$ of the cube's corners  we proceed as follows.
First, starting from a cropped version of the LiDAR pointcloud (Figure \ref{fig:lid_pipeline} (a)), we  fit 3 planes $\pi_1, \pi_2, \pi_3$ using Sequential Ransac (Figure \ref{fig:lid_pipeline} (b)).   
Then  $E_0$ is computed as the intersection of all the three planes $\pi_1$, $\pi_2$, and $\pi_3$. Since the dimensions of the cube are known by design, the remaining six points $\{E_1, E_2, ..., E_6\}$ are derived through standard geometric computations, as illustrated in Figure~\ref{fig:lid_pipeline}   (c).

\subsection{ArUco markers detection in RGB and pointcloud}
\label{sec:feat_RGB}
Starting from the cube's corners $\{E_i\}$ identified in the pointcloud, the ArUco markers $\{A_i\}$ are identified exploiting the known measures of each cube's face. In this way we can identify 4 points (the corners) for each ArUco marker in the pointcloud, for a total of 60 points $\{A_0,A_1,...,A_{59}\}$. The corresponding  markers $\{a_0,a_1,...,a_{59}\}$ are retrieved from the RGB camera frames using the marker detection procedure from Garrido-Juardo \etal \cite{garrido2014automatic}.

\subsection{Pose estimation}
\label{sec:pose_est}
For both calibrations, the one between LiDAR and event camera and the one between LiDAR and RGB camera, the calibration procedure is made by employing the well-known Perspective-n-points algorithm \cite{zheng2013revisiting} applied on the correspondences found. In particular, for the calibration between the LiDAR and the event camera we use the 7 cube's corners identified in both the  event camera frame and in the pointcloud, building a set of correspondences $\{E_i\leftrightarrow e_i\}$ with $i=0,\ldots,6$.  For the calibration between the LiDAR and the RGB camera we use the corners of the ArUco markers identified in both the camera image plane and in the pointcloud, building the set of correspondences $\{A_i\leftrightarrow a_i\}$ with $i=0,\ldots, 59$.

\section{Experimental validation}

All the experiments are conducted on a custom recorded dataset composed of 17 sequences featuring our calibration target in different light conditions (morning, noon and evening) and positions. The dataset has been recorded on a Maserati GranCabrio Folgore equipped with three LiDARs, three RGB cameras and one event camera, all directly calibrated with our method (Figure \ref{fig:cube}). 

\begin{figure}
\centering
\begin{tabular}{ccc}
\includegraphics[height=3.5cm]{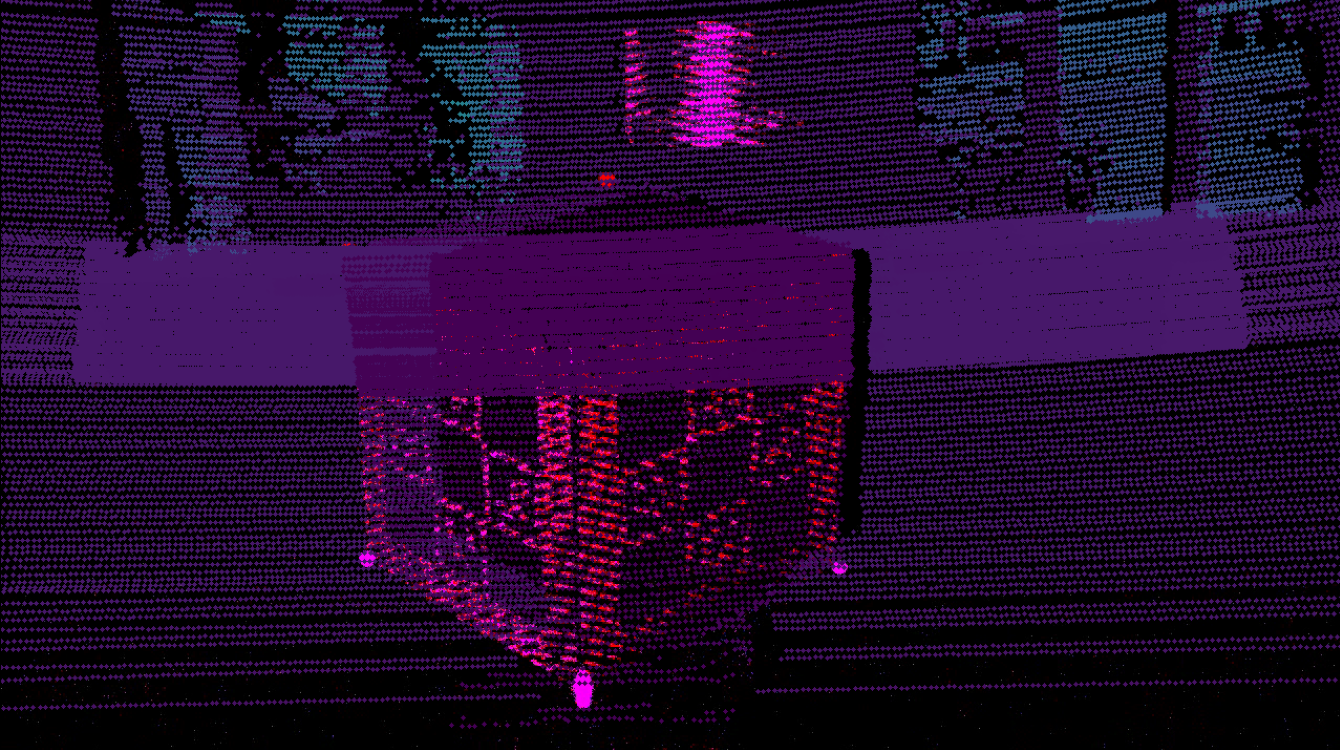} &
\includegraphics[height=3.5cm]{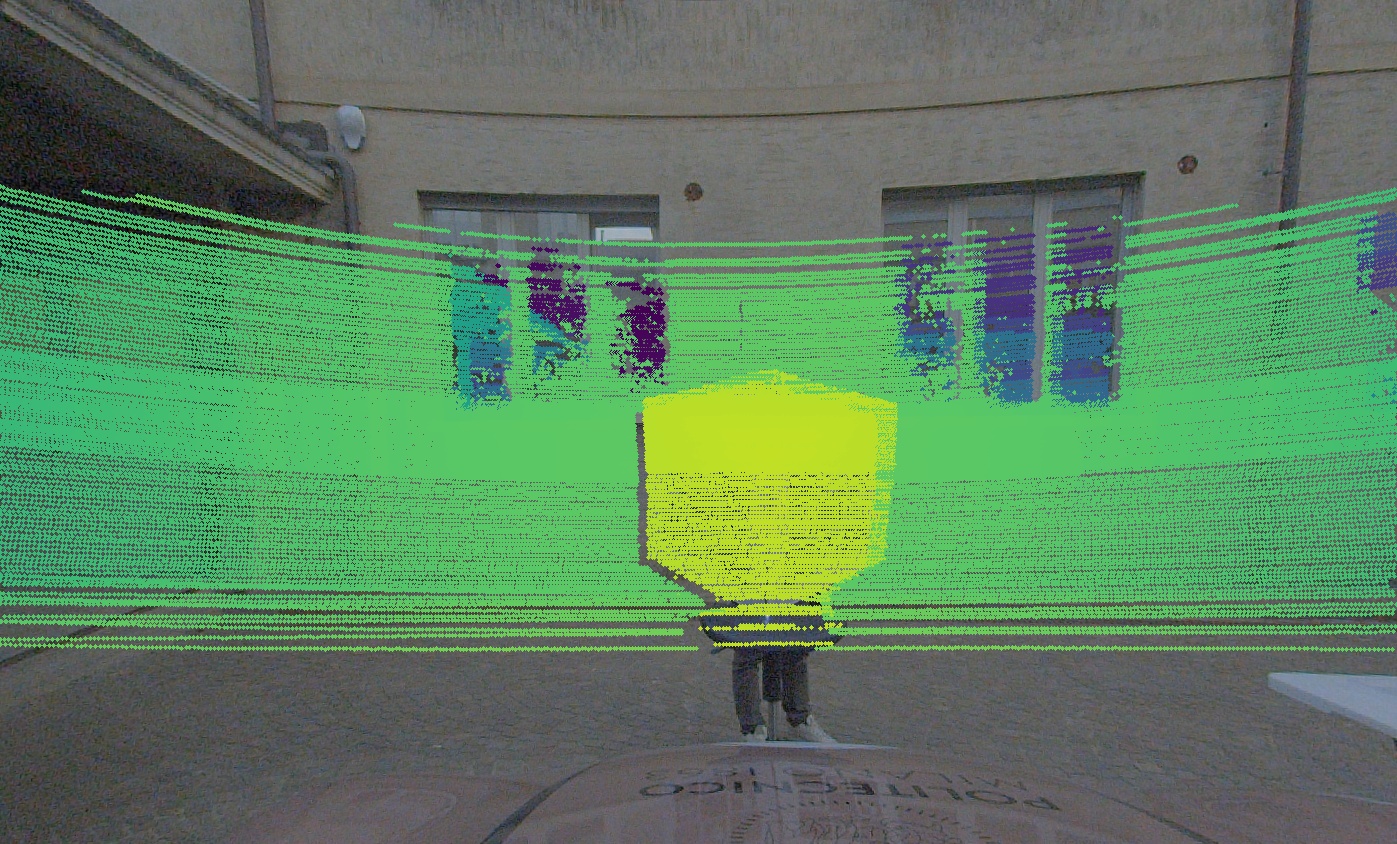}\\
(a) & (b)
\end{tabular}
\caption{Qualitative analysis results. The alignment between the pointcloud and the  event stream  in (a) and the alignement between the  pointcloud and the RGB image (b) are satisfactory. From (a) can also be noticed that the cube's faces are illuminated by the LiDAR and the reflection triggers events.
}
\label{fig:projection}
\end{figure}

\paragraph{Figure of merits.}
We quantitatively evaluate  calibration performances in terms of \emph{average reprojection error}
   $ E_{\text{mean}} = \frac{1}{n} \sum_{i=1}^{n} \sqrt{(x_i - \hat{x}_i)^2 + (y_i - \hat{y}_i)^2}
$, where the reprojection error measures the image distance in pixels between a projected point $(x, y)$ and a measured one $(\hat{x}, \hat{y})$. We consider the average error between detected points in the target reference frame $j$ and the projection of the same points projected from the source reference frame $i$ to the target $j$ by means of the projection matrix $P_{i,j}=[R_{i,j} | t_{i,j}]$. The errors are also averaged by calibration type, yielding one value for the event camera–LiDAR calibration and another for the RGB–LiDAR calibration. This is necessary since multiple LiDARs are calibrated with the event camera and multiple LiDARs are calibrated with different RGB cameras.
To qualitatively evaluate the calibrations involving LiDARs, the pointcloud is projected onto the image plane (Figure \ref{fig:projection}) and the alignment between the two is evaluated.

\paragraph{Alternative methods.}
In terms of performances, we can compare our method against those that use the reprojection error as performance measure. In fact, since we used a custom calibration target, we can not evaluate our method on standard datasets provided with ground truth.
As regards the calibration between the event camera and the LiDARs we consider Song \etal \cite{song2018calibration} that uses a custom calibration target and Xing \etal \cite{xing2023target} that instead relies on  a targetless calibration approach. 
As regards the calibration between RGB cameras and LiDAR, we consider  Yan \etal \cite{yan2023joint}, Grammatikopoulos \etal \cite{grammatikopoulos2022effective} and  Zhou \etal \cite{zhou2018automatic} which all exploit different calibration targets.

\paragraph{Results.} Results are collated in  Tab \ref{tab:table_res}. For what concerns the more challenging calibration between the event camera and the LiDAR, the benefits of our multi-modal calibration target are more evident, as our performances are considerably better than the ones obtained by similar methods. As regards the calibration between the RGB camera and the LiDAR the results are comparable, the difference with the best method is only 0.03 pixels, which is indeed a very small deviation. The advantage of our approach compared to the others is that we can calibrate the LiDAR with both the event and the RGB camera with a single calibration target and with a unique pipeline, in contrast to previously published methods which take into account only one type of calibration. 

\begin{table}
\centering
\resizebox{0.9\textwidth}{!}{%
\begin{tabular}{|c|c|c|c|}
\hline
Method & Event  camera - LiDAR & RGB  Camera - LiDAR\\
\hline
Song \etal \cite{song2018calibration} & 4.691 & -  \\
Xing \etal \cite{xing2023target} & 3.161 & -  \\
Yan \etal \cite{yan2023joint} & - & 1.104 \\
Grammatikopoulos \etal \cite{grammatikopoulos2022effective} & - & 11.4826 *\\
Zhou \etal\cite{zhou2018automatic} & - & 21.69332 *\\
\textbf{Our} & \textbf{2.732} & \textbf{1.134}  \\
\hline
\end{tabular}
}
\caption{The quantitative results (in pixels) compared with other similar works on the same metric. Our method is the only one that takes into account both the calibrations directly. 
An asterisk * indicates values computed from the vertical reprojection error and the horizontal reprojection error reported by the authors.}
\label{tab:table_res}
\end{table}

\paragraph{The impact of a geometric correction.}
As explained in Section \ref{sec:corners_events} we assume that the centers $\{e_i\}$  of the ellipses coincide with the projections of the corners of the cube $\{E_i\}$. A more sophisticated approximation is to treat  each LED diode as a 3D ellipsoid, and to consider its projection as an ellipse in the image plane. Given at least two ellipses-ellipsoids correspondences (we have seven) we can estimate the event camera pose by exploiting the algorithm proposed by Gaudilliere \etal \cite{gaudilliere2020perspective} that leverages the geometric properties of quadric projection to accurately localize a perspective camera. 
The performances of this refinement applied to a synthetic scene (where the quadric definition of each ellipsoid is correct) are good (0.3 of average Jaccard distance between detected ellipses and projected ellipsoids) but even if this method is theoretically correct, the accuracy drops in real world scenes where we have to estimate the ellipsoidal model of a LED, therefore we decided to simply approximate the projections of the corners with the ellipses' centers.

\section{Practical implementations}
The work proposed has been implemented  to calibrate a real autonomous driving framework of a fully autonomous Maserati GrancCabrio Folgore. Further details are available at \href{https://andberto.github.io/One-target-to-align-them-all/}{https://andberto.github.io/One-target-to-align-them-all/}.
The sensors suite is adapted from \cite{sgaravatti2024multimodal, pieroni2024multi} with the addition of the event cameras.
As introduced in Figure \ref{fig:cube}  the sensors employed are:
\begin{itemize}[nosep] 
    \item[\blackletter{A}] Two \href{https://ouster.com/products/hardware/os1-lidar-sensor}{Ouster LiDARs 360° Mid Range}, which capture a dense 3D representation of the surroundings (128 scan-lines) and a distance range up to 200m.
    \item[\blackletter{B}] One \href{https://www.seyond.com/products/falcon-k1/}{Seyond Falcon K Ultra-Long Range} which, with a 120° horizontal FoV, samples distances up to 500m on 152 scan-lines.
    \item[\blackletter{C}] Up to three \href{https://thinklucid.com/ip67-cameras-and-accessories/}{Lucid Vision IP68 RGB Cameras}.
    \item[\blackletter{D}] One \href{https://www.prophesee.ai/event-camera-evk4/}{EVK4 Prophesee} provided with the IMX636 (HD) sensor.
\end{itemize}
The data is exchanged with the elaboration system through the well-known ROS2 Macenski \etal \cite{macenski2022robot}.

\section{Conclusions}
In this work, we successfully addressed the challenge of event camera calibration in multi-sensor autonomous driving systems. Our novel calibration target and feature extraction pipeline specifically tackle the difficulties inherent in event-based sensing, while simultaneously enabling standard RGB and LiDAR calibration within a unified framework.

Our approach lends itself well to extensions and adaptations, in fact the calibration pipeline could be extended to also include a stereo calibration between the two cameras following an approach similar to the one presented by Garcia \etal \cite{garcia2025marker}. The calibration marker could be extended to include more feature points by adding more LEDs in order to increase the accuracy of the PnP algorithm. In general, our approach lays the foundation for a wide range of applications that require a robust, accurate, and flexible extrinsic calibration as a starting point for complex sensor systems, as for example multi-modal autonomous driving object detection frameworks.

\subsubsection*{Acknowledgments}
The authors gratefully acknowledge Riccardo Pieroni, Andrea Corno, and Matteo Savaresi, along with the AIDA team, for their invaluable support and for generously providing access to their autonomous driving suite, vehicle, and calibration framework.

\bibliographystyle{unsrtnat}
\bibliography{references}  

@inproceedings{ta2023l2e,
  title={L2e: Lasers to events for 6-dof extrinsic calibration of lidars and event cameras},
  author={Ta, Kevin and Bruggemann, David and Br{\"o}dermann, Tim and Sakaridis, Christos and Van Gool, Luc},
  booktitle={2023 IEEE International Conference on Robotics and Automation (ICRA)},
  pages={11425--11431},
  year={2023},
  organization={IEEE}
}

@article{jiao2023lce,
  title={LCE-Calib: automatic lidar-frame/event camera extrinsic calibration with a globally optimal solution},
  author={Jiao, Jianhao and Chen, Feiyi and Wei, Hexiang and Wu, Jin and Liu, Ming},
  journal={IEEE/ASME Transactions on Mechatronics},
  volume={28},
  number={5},
  pages={2988--2999},
  year={2023},
  publisher={IEEE}
}

@article{rebecq2019high,
  title={High speed and high dynamic range video with an event camera},
  author={Rebecq, Henri and Ranftl, Ren{\'e} and Koltun, Vladlen and Scaramuzza, Davide},
  journal={IEEE transactions on pattern analysis and machine intelligence},
  volume={43},
  number={6},
  pages={1964--1980},
  year={2019},
  publisher={IEEE}
}

@article{xing2023target,
  title={Target-free extrinsic calibration of event-lidar dyad using edge correspondences},
  author={Xing, Wanli and Lin, Shijie and Yang, Lei and Pan, Jia},
  journal={IEEE Robotics and Automation Letters},
  volume={8},
  number={7},
  pages={4020--4027},
  year={2023},
  publisher={IEEE}
}

@inproceedings{song2018calibration,
  title={Calibration of event-based camera and 3d lidar},
  author={Song, Rihui and Jiang, Zhihua and Li, Yanghao and Shan, Yunxiao and Huang, Kai},
  booktitle={2018 WRC Symposium on Advanced Robotics and Automation (WRC SARA)},
  pages={289--295},
  year={2018},
  organization={IEEE}
}

@article{li2023automatic,
  title={Automatic targetless LiDAR--camera calibration: a survey},
  author={Li, Xingchen and Xiao, Yuxuan and Wang, Beibei and Ren, Haojie and Zhang, Yanyong and Ji, Jianmin},
  journal={Artificial Intelligence Review},
  volume={56},
  number={9},
  pages={9949--9987},
  year={2023},
  publisher={Springer}
}

@ARTICLE{survey2,

  author={An, Pei and Ding, Junfeng and Quan, Siwen and Yang, Jiaqi and Yang, You and Liu, Qiong and Ma, Jie},

  journal={IEEE Transactions on Intelligent Transportation Systems}, 

  title={Survey of Extrinsic Calibration on LiDAR-Camera System for Intelligent Vehicle: Challenges, Approaches, and Trends}, 

  year={2024},

  volume={25},

  number={11},

  pages={15342-15366},

  doi={10.1109/TITS.2024.3419758}}

@inproceedings{geiger2012automatic,
  title={Automatic camera and range sensor calibration using a single shot},
  author={Geiger, Andreas and Moosmann, Frank and Car, {\"O}mer and Schuster, Bernhard},
  booktitle={2012 IEEE international conference on robotics and automation},
  pages={3936--3943},
  year={2012},
  organization={IEEE}
}

@article{gehrig2021dsec,
  title={Dsec: A stereo event camera dataset for driving scenarios},
  author={Gehrig, Mathias and Aarents, Willem and Gehrig, Daniel and Scaramuzza, Davide},
  journal={IEEE Robotics and Automation Letters},
  volume={6},
  number={3},
  pages={4947--4954},
  year={2021},
  publisher={IEEE}
}

@article{zhu2018multivehicle,
  title={The multivehicle stereo event camera dataset: An event camera dataset for 3D perception},
  author={Zhu, Alex Zihao and Thakur, Dinesh and {\"O}zaslan, Tolga and Pfrommer, Bernd and Kumar, Vijay and Daniilidis, Kostas},
  journal={IEEE Robotics and Automation Letters},
  volume={3},
  number={3},
  pages={2032--2039},
  year={2018},
  publisher={IEEE}
}

@inproceedings{zhang2004extrinsic,
  title={Extrinsic calibration of a camera and laser range finder (improves camera calibration)},
  author={Zhang, Qilong and Pless, Robert},
  booktitle={2004 IEEE/RSJ International Conference on Intelligent Robots and Systems (IROS)(IEEE Cat. No. 04CH37566)},
  volume={3},
  pages={2301--2306},
  year={2004},
  organization={IEEE}
}

@inproceedings{zhou2012new,
  title={A new algorithm for computing the projection matrix between a LIDAR and a camera based on line correspondences},
  author={Zhou, Lipu and Deng, Zhidong},
  booktitle={2012 IV International Congress on Ultra Modern Telecommunications and Control Systems},
  pages={436--441},
  year={2012},
  organization={IEEE}
}

@article{dhall2017lidar,
  title={LiDAR-camera calibration using 3D-3D point correspondences},
  author={Dhall, Ankit and Chelani, Kunal and Radhakrishnan, Vishnu and Krishna, K Madhava},
  journal={arXiv preprint arXiv:1705.09785},
  year={2017}
}

@article{yoo2018improved,
  title={Improved LiDAR-camera calibration using marker detection based on 3D plane extraction},
  author={Yoo, Joong-Sun and Kim, Do-Hyeong and Kim, Gon-Woo},
  journal={Journal of Electrical Engineering and Technology},
  volume={13},
  number={6},
  pages={2530--2544},
  year={2018},
  publisher={The Korean Institute of Electrical Engineers}
}

@inproceedings{guindel2017automatic,
  title={Automatic extrinsic calibration for lidar-stereo vehicle sensor setups},
  author={Guindel, Carlos and Beltr{\'a}n, Jorge and Mart{\'\i}n, David and Garc{\'\i}a, Fernando},
  booktitle={2017 IEEE 20th international conference on intelligent transportation systems (ITSC)},
  pages={1--6},
  year={2017},
  organization={IEEE}
}

@inproceedings{velas2014calibration,
  title={Calibration of RGB camera with Velodyne LiDAR},
  author={Vel{\'a}s, Martin and {\v{S}}pan{\v{e}}l, Michal and Materna, Zden{\v{e}}k and Herout, Adam},
  booktitle={WSCG 2014 Conference on Computer Graphics, Visualization and Computer Vision},
  year={2014},
  publisher={V{\'a}clav Skala - UNION Agency},
  address={Plzeň, Czech Republic}
}

@article{hassanein2016new,
  title={A new automatic system calibration of multi-cameras and lidar sensors},
  author={Hassanein, M and Moussa, A and El-Sheimy, N},
  journal={The International Archives of the Photogrammetry, Remote Sensing and Spatial Information Sciences},
  volume={41},
  pages={589--594},
  year={2016},
  publisher={Copernicus GmbH}
}

@inproceedings{pusztai2017accurate,
  title={Accurate calibration of LiDAR-camera systems using ordinary boxes},
  author={Pusztai, Zoltan and Hajder, Levente},
  booktitle={Proceedings of the IEEE international conference on computer vision workshops},
  pages={394--402},
  year={2017}
}

@inproceedings{toth2020automatic,
  title={Automatic LiDAR-camera calibration of extrinsic parameters using a spherical target},
  author={T{\'o}th, Tekla and Pusztai, Zolt{\'a}n and Hajder, Levente},
  booktitle={2020 IEEE International Conference on Robotics and Automation (ICRA)},
  pages={8580--8586},
  year={2020},
  organization={IEEE}
}

@article{pervsic2021online,
  title={Online multi-sensor calibration based on moving object tracking},
  author={Per{\v{s}}i{\'c}, Juraj and Petrovi{\'c}, Luka and Markovi{\'c}, Ivan and Petrovi{\'c}, Ivan},
  journal={Advanced Robotics},
  volume={35},
  number={3-4},
  pages={130--140},
  year={2021},
  publisher={Taylor \& Francis}
}

@inproceedings{li2017online,
  title={Online high-accurate calibration of RGB+ 3D-LiDAR for autonomous driving},
  author={Li, Tao and Fang, Jianwu and Zhong, Yang and Wang, Di and Xue, Jianru},
  booktitle={Image and Graphics: 9th International Conference, ICIG 2017, Shanghai, China, September 13-15, 2017, Revised Selected Papers, Part III 9},
  pages={254--263},
  year={2017},
  organization={Springer}
}

@inproceedings{liu2018deep,
  title={A deep-learning based multi-modality sensor calibration method for usv},
  author={Liu, Hao and Liu, Yingjian and Gu, Xiaoyan and Wu, Yingying and Qu, Fangchao and Huang, Lei},
  booktitle={2018 IEEE Fourth International Conference on Multimedia Big Data (BigMM)},
  pages={1--5},
  year={2018},
  organization={IEEE}
}

@article{beltran2022automatic,
  title={Automatic extrinsic calibration method for lidar and camera sensor setups},
  author={Beltr{\'a}n, Jorge and Guindel, Carlos and De La Escalera, Arturo and Garc{\'\i}a, Fernando},
  journal={IEEE Transactions on Intelligent Transportation Systems},
  volume={23},
  number={10},
  pages={17677--17689},
  year={2022},
  publisher={IEEE}
}

@article{yuan2021pixel,
  title={Pixel-level extrinsic self calibration of high resolution lidar and camera in targetless environments},
  author={Yuan, Chongjian and Liu, Xiyuan and Hong, Xiaoping and Zhang, Fu},
  journal={IEEE Robotics and Automation Letters},
  volume={6},
  number={4},
  pages={7517--7524},
  year={2021},
  publisher={IEEE}
}

@article{zhao2024power,
  title={Power equipment vibration visualization using intelligent sensing method based on event-sensing principle},
  author={M. Zhao and X. Shen and L. Su and Z. Dong},
  journal={Global Energy Interconnection},
  volume={7},
  number={2},
  pages={228--240},
  year={2024},
  publisher={Elsevier}
}

@article{fitzgibbon1999direct,
  title={Direct least square fitting of ellipses},
  author={Fitzgibbon, Andrew and Pilu, Maurizio and Fisher, Robert B},
  journal={IEEE Transactions on pattern analysis and machine intelligence},
  volume={21},
  number={5},
  pages={476--480},
  year={1999},
  publisher={IEEE}
}

@article{garrido2014automatic,
  title={Automatic generation and detection of highly reliable fiducial markers under occlusion},
  author={Garrido-Jurado, Sergio and Mu{\~n}oz-Salinas, Rafael and Madrid-Cuevas, Francisco Jos{\'e} and Mar{\'\i}n-Jim{\'e}nez, Manuel Jes{\'u}s},
  journal={Pattern Recognition},
  volume={47},
  number={6},
  pages={2280--2292},
  year={2014},
  publisher={Elsevier}
}

@inproceedings{yan2023joint,
  title={Joint camera intrinsic and lidar-camera extrinsic calibration},
  author={Yan, Guohang and He, Feiyu and Shi, Chunlei and Wei, Pengjin and Cai, Xinyu and Li, Yikang},
  booktitle={2023 IEEE International Conference on Robotics and Automation (ICRA)},
  pages={11446--11452},
  year={2023},
  organization={IEEE}
}

@article{gaudilliere2020perspective,
  title={Perspective-2-ellipsoid: Bridging the gap between object detections and 6-dof camera pose},
  author={Gaudilli{\`e}re, Vincent and Simon, Gilles and Berger, Marie-Odile},
  journal={IEEE Robotics and Automation Letters},
  volume={5},
  number={4},
  pages={5189--5196},
  year={2020},
  publisher={IEEE}
}

@inproceedings{censi2014low,
  title={Low-latency event-based visual odometry},
  author={Censi, Andrea and Scaramuzza, Davide},
  booktitle={2014 IEEE International Conference on Robotics and Automation (ICRA)},
  pages={703--710},
  year={2014},
  organization={IEEE}
}

@inproceedings{zuo2019lic,
  title={Lic-fusion: Lidar-inertial-camera odometry},
  author={Zuo, Xingxing and Geneva, Patrick and Lee, Woosik and Liu, Yong and Huang, Guoquan},
  booktitle={2019 IEEE/RSJ International Conference on Intelligent Robots and Systems (IROS)},
  pages={5848--5854},
  year={2019},
  organization={IEEE}
}

@article{magrini2024neuromorphic,
  title={Neuromorphic Drone Detection: an Event-RGB Multimodal Approach},
  author={Magrini, Gabriele and Becattini, Federico and Pala, Pietro and Del Bimbo, Alberto and Porta, Antonio},
  journal={arXiv preprint arXiv:2409.16099},
  year={2024}
}

@inproceedings{wu2023flytracker,
  title={Flytracker: Motion tracking and obstacle detection for drones using event cameras},
  author={Wu, Yue and Xu, Jingao and Li, Danyang and Xie, Yadong and Cao, Hao and Li, Fan and Yang, Zheng},
  booktitle={IEEE INFOCOM 2023-IEEE Conference on Computer Communications},
  pages={1--10},
  year={2023},
  organization={IEEE}
}

@article{sgaravatti2025multimodal,
  title={A Multimodal Hybrid Late-Cascade Fusion Network for Enhanced 3D Object Detection},
  author={Sgaravatti, Carlo and Basla, Roberto and Pieroni, Riccardo and Corno, Matteo and Savaresi, Sergio M and Magri, Luca and Boracchi, Giacomo},
  journal={arXiv preprint arXiv:2504.18419},
  year={2025}
}

@inproceedings{bai2022transfusion,
  title={Transfusion: Robust lidar-camera fusion for 3d object detection with transformers},
  author={Bai, Xuyang and Hu, Zeyu and Zhu, Xinge and Huang, Qingqiu and Chen, Yilun and Fu, Hongbo and Tai, Chiew-Lan},
  booktitle={Proceedings of the IEEE/CVF conference on computer vision and pattern recognition},
  pages={1090--1099},
  year={2022}
}

@inproceedings{yin2023fgfusion,
  title={Fgfusion: Fine-grained lidar-camera fusion for 3d object detection},
  author={Yin, Zixuan and Sun, Han and Liu, Ningzhong and Zhou, Huiyu and Shen, Jiaquan},
  booktitle={Chinese Conference on Pattern Recognition and Computer Vision (PRCV)},
  pages={505--517},
  year={2023},
  organization={Springer}
}

@article{pfrommer2022frequency,
  title={Frequency cam: Imaging periodic signals in real-time},
  author={Pfrommer, Bernd},
  journal={arXiv preprint arXiv:2211.00198},
  year={2022}
}

@inproceedings{hoseini2017passive,
  title={Passive localization and detection of quadcopter UAVs by using dynamic vision sensor},
  author={Hoseini, Sahar and Orchard, Garrick and Yousefzadeh, Amirreza and Deverakonda, Balakrishna and Serrano-Gotarredona, Teresa and Linares-Barranco, Bernab{\'e}},
  booktitle={2017 5th Iranian Joint Congress on Fuzzy and Intelligent Systems (CFIS)},
  pages={81--85},
  year={2017},
  organization={IEEE}
}

@inproceedings{censi2013low,
  title={Low-latency localization by active LED markers tracking using a dynamic vision sensor},
  author={Censi, Andrea and Strubel, Jonas and Brandli, Christian and Delbruck, Tobi and Scaramuzza, Davide},
  booktitle={2013 IEEE/RSJ International Conference on Intelligent Robots and Systems},
  pages={891--898},
  year={2013},
  organization={IEEE}
}

@article{grammatikopoulos2022effective,
  title={An effective camera-to-lidar spatiotemporal calibration based on a simple calibration target},
  author={Grammatikopoulos, Lazaros and Papanagnou, Anastasios and Venianakis, Antonios and Kalisperakis, Ilias and Stentoumis, Christos},
  journal={Sensors},
  volume={22},
  number={15},
  pages={5576},
  year={2022},
  publisher={MDPI}
}

@inproceedings{zhou2018automatic,
  title={Automatic extrinsic calibration of a camera and a 3d lidar using line and plane correspondences},
  author={Zhou, Lipu and Li, Zimo and Kaess, Michael},
  booktitle={2018 IEEE/RSJ International Conference on Intelligent Robots and Systems (IROS)},
  pages={5562--5569},
  year={2018},
  organization={IEEE}
}

@ARTICLE{canny,
  author={Canny, John},
  journal={IEEE Transactions on Pattern Analysis and Machine Intelligence}, 
  title={A Computational Approach to Edge Detection}, 
  year={1986},
  volume={PAMI-8},
  number={6},
  pages={679-698},
  doi={10.1109/TPAMI.1986.4767851}}

@article{garcia2025marker,
  title={Marker-Based Extrinsic Calibration Method for Accurate Multi-Camera 3D Reconstruction},
  author={Garcia-D'Urso, Nahuel and Sanchez-Sos, Bernabe and Azorin-Lopez, Jorge and Fuster-Guillo, Andres and Macia-Lillo, Antonio and Mora-Mora, Higinio},
  journal={arXiv preprint arXiv:2505.02539},
  year={2025}
}

@inproceedings{zheng2013revisiting,
  title={Revisiting the pnp problem: A fast, general and optimal solution},
  author={Zheng, Yinqiang and Kuang, Yubin and Sugimoto, Shigeki and Astrom, Kalle and Okutomi, Masatoshi},
  booktitle={Proceedings of the IEEE International Conference on Computer Vision},
  pages={2344--2351},
  year={2013}
}

@article{hu2024dynamic,
  title={A Dynamic Calibration Framework for the Event-Frame Stereo Camera System},
  author={Hu, Rui and Kogler, J{\"u}rgen and Gelautz, Margrit and Lin, Min and Xia, Yuanqing},
  journal={IEEE Robotics and Automation Letters},
  year={2024},
  publisher={IEEE}
}

@inproceedings{dubeau2020rgb,
  title={RGB-DE: Event camera calibration for fast 6-dof object tracking},
  author={Dubeau, Etienne and Garon, Mathieu and Debaque, Benoit and de Charette, Raoul and Lalonde, Jean-Fran{\c{c}}ois},
  booktitle={2020 IEEE International Symposium on Mixed and Augmented Reality (ISMAR)},
  pages={127--135},
  year={2020},
  organization={IEEE}
}

@article{cress2024tumtraf,
  title={Tumtraf event: Calibration and fusion resulting in a dataset for roadside event-based and rgb cameras},
  author={Cre{\ss}, Christian and Zimmer, Walter and Purschke, Nils and Doan, Bach Ngoc and Kirchner, Sven and Lakshminarasimhan, Venkatnarayanan and Strand, Leah and Knoll, Alois C},
  journal={IEEE Transactions on Intelligent Vehicles},
  year={2024},
  publisher={IEEE}
}

@inproceedings{muglikar2021calibrate,
  title={How to calibrate your event camera},
  author={Muglikar, Manasi and Gehrig, Mathias and Gehrig, Daniel and Scaramuzza, Davide},
  booktitle={Proceedings of the IEEE/CVF conference on computer vision and pattern recognition},
  pages={1403--1409},
  year={2021}
}

@article{macenski2022robot,
  title={Robot operating system 2: Design, architecture, and uses in the wild},
  author={Macenski, Steven and Foote, Tully and Gerkey, Brian and Lalancette, Chris and Woodall, William},
  journal={Science robotics},
  volume={7},
  number={66},
  pages={eabm6074},
  year={2022},
  publisher={American Association for the Advancement of Science}
}

@inproceedings{sgaravatti2024multimodal,
  title={A multimodal hybrid late-cascade fusion network for enhanced 3d object detection},
  author={Sgaravatti, Carlo and Basla, Roberto and Pieroni, Riccardo and Corno, Matteo and Savaresi, Sergio M and Magri, Luca and Boracchi, Giacomo},
  booktitle={European Conference on Computer Vision},
  pages={339--356},
  year={2024},
  organization={Springer}
}

@inproceedings{pieroni2024multi,
  title={Multi-object tracking with camera-LiDAR fusion for autonomous driving},
  author={Pieroni, Riccardo and Specchia, Simone and Corno, Matteo and Savaresi, Sergio Matteo},
  booktitle={2024 European Control Conference (ECC)},
  pages={2774--2779},
  year={2024},
  organization={IEEE}
}






\end{document}